\documentclass{article}

 \usepackage[preprint]{neurips_2019}

\usepackage{natbib}
\usepackage{graphicx}
\usepackage[utf8]{inputenc} % allow utf-8 input
\usepackage[T1]{fontenc}    % use 8-bit T1 fonts
\usepackage{hyperref}       % hyperlinks
\usepackage{url}            % simple URL typesetting
\usepackage{booktabs}       % professional-quality tables
\usepackage{amsfonts}       % blackboard math symbols
\usepackage{nicefrac}       % compact symbols for 1/2, etc.
\usepackage{microtype}      % microtypography
\usepackage{times}
\usepackage{latexsym}
\usepackage{multirow}
\usepackage{color}
\usepackage{graphicx}
\usepackage{enumitem}
\usepackage{CJKutf8}
\usepackage{comment}
\usepackage[justification=centering]{caption}
\usepackage{amsmath}

  {\begin{itemize}[topsep=0pt, partopsep=0pt] %
    \setlength{\itemsep}{0pt}%
    \setlength{\parskip}{0pt}%
    }%
  {\end{itemize}}
  \usepackage[font=small]{caption}  % !
\usepackage[font=small]{subcaption}  % !
  \usepackage{color}
  {\begin{enumerate}â \footnotesize%
    \setlength{\itemsep}{0pt}%
    \setlength{\parskip}{0pt}%
    }%
  {\end{enumerate}}

\usepackage{url}

\usepackage{url}

\title{Large-scale Pretraining for Neural Machine Translation with Tens of Billions of Sentence Pairs}

 \usepackage[preprint]{neurips_2019}

\author{
Yuxian Meng$^{1*}$,  Xiangyuan Ren$^{1*}$,  Zijun Sun$^{1*}$, \\
\bf{ Xiaoya Li$^1$, Arianna Yuan$^2$, Fei Wu$^3$ and Jiwei Li$^1$}\\
\\
$^1$ ShannonAI\\
$^2$ Computer Science Department, Stanford University\\
$^3$ Department of Computer Science and Technology, Zhejiang University\\}

\usepackage{times}
\usepackage{latexsym}
\usepackage{multirow}
\usepackage{comment}

\usepackage{color}
\usepackage{graphicx}

\usepackage{times}
\usepackage{url}
\usepackage{latexsym}
\usepackage{xspace}
\usepackage{booktabs}
\usepackage{color}
\usepackage{graphicx}
\usepackage{tabulary}
\usepackage{enumitem}
\usepackage{amsfonts}

\usepackage{microtype}
\usepackage{multirow}
\usepackage{verbatim}
\usepackage{amsmath,amsthm,amssymb}
\usepackage{array}
\usepackage[scaled=0.86]{helvet}
\usepackage{ifthen}
\usepackage{courier}
\usepackage[linesnumbered,vlined,ruled]{algorithm2e}
 \usepackage{lipsum}
\usepackage[export]{adjustbox}

\newcommand{\captionfonts}{\small}
\makeatletter  % Allow the use of @ in command names
\long\def\@makecaption#1#2{%
  \vskip\abovecaptionskip
  \sbox\@tempboxa{{\captionfonts #1: #2}}%
  \ifdim \wd\@tempboxa >\hsize
    {\captionfonts #1: #2\par}
  \else
    \hbox to\hsize{\hfil\box\@tempboxa\hfil}%
  \fi
  \vskip\belowcaptionskip}
\makeatother   % Cancel the effect of \makeatletter
\usepackage{soul}

\belowdisplayskip \abovedisplayskip

\usepackage{booktabs}
\usepackage{subcaption}
\usepackage{lipsum}

\newcommand{\sts}{{{\textsc{Seq2Seq}}}\xspace}

\usepackage{CJKutf8}

\begin{document}
\begin{CJK*}{UTF8}{gbsn}

\maketitle

\begin{abstract}
In this paper, we  investigate the problem of training neural machine translation (NMT) 
systems
with a dataset of more than 40 billion  bilingual sentence pairs, which is larger than the largest dataset to date by orders of magnitude. 
Unprecedented challenges emerge in this situation  compared to  previous NMT work, including severe noise in the data and prohibitively long training time. We propose practical solutions to handle these issues and demonstrate that large-scale pretraining significantly improves NMT performance. 
We are able to push the
 BLEU score of  WMT17  Chinese-English dataset to 32.3, with a significant performance boost of +3.2 over existing state-of-the-art results.
% \footnote{Yuxian, Xiangyuan and Zijun contributed equally to this paper.}
% \footnote{
% \texttt{\{yuxian\_meng, xiangyuan\_ren, zijun\_sun and jiwei\_li\} @shannonai.com},
% \texttt{yaowen@nudt.edu.cn},
% \texttt{junzhang.1227@outlook.com},
% \texttt{wufei@zju.edu.cn}
% \texttt{xfyuan@stanford.edu}}
\end{abstract}

\section{Introduction}
End-to-end neural machine translation (NMT) \citep{bahdanau2014neural,sutskever2014sequence,luong2015effective,sennrich2015neural,vaswani2017attention,britz2017massive,gehring2017convolutional,klein2017opennmt,johnson2017google,wang2017sogou,hassan2018achieving,aharoni2019massively,ng2019facebook}
 has been widely adopted as the state-of-the-art approach for MT. 
Particularly, 
sequence-to-sequence models (\sts for short) are trained and tested on
 publicly 
 available
 benchmark datasets, the size of which ranges from tens of thousands  for  low-resource languages 
to hundreds of millions for widely used languages.  

Recent progress in natural language understanding (NLU) has proved that large-scale pretraining  
such
as BERT \citep{devlin2018bert} or Elmo \citep{peters2018deep} often lead to a significant leap forward in SOTA results. 
Mostly  due to the lack of paired training data \citep{song2019mass},
no comparable success has been made in the field of MT, 
Therefore, two important  questions remained unanswered:
\textsc{whether}  we  can push the performance of existing neural models with more data, e.g., if we use tens of  billions of bilingual training sentence pairs. And if so, \textsc{how} we can address the unique challenges introduced by such gigantic dataset. 
The answers to these questions are not immediately clear as several issues are unprecedented in this case:
\begin{itemize}
\item Scale:  
Firstly, an NMT model's expressivity 
 is limited by  infrastructures such as  GPU memory, so indefinitely increasing the size of the training data might not improve the performance. 
Secondly, training on a massively large dataset with tens of billion sentence pairs can be prohibitively slow.

\item Noise and Out-of-domain data: 
A dataset with tens of billions of 
bilingual sentences pairs
must  span a wide range of domains and comes from extremely noisy sources. 
It is widely accepted  that
 translation quality 
 is very vulnerable to out-of-domain data and noisy data \citep{chen2016guided,niehues2010domain,koehn2007experiments,eidelman2012topic}, and a small number of noisy training instances can have negative effects on translation quality \citep{belinkov2017synthetic}. 
     This means blindly increasing the size of training data
   or adding noisy training data may
 not necessarily lead to a better performance, but may  even backfire. 
\end{itemize}

In this paper, we  investigate the problem of training neural machine translation  
systems
on a dataset with more than 40 billion  bilingual sentence pairs, which is by orders of magnitude larger than the largest dataset to date. 
To tailor existing WMT systems to the massive training dataset, we propose
a pipelined strategy which involves  large-scale pretraining and  domain-specific fine-tuning.
 We handle  the trade-off between 
 full-dataset optimality and convergence speed for pretraining and
 demonstrate that having
 large-scale pretraining  significantly improves NMT performance.
 
Combining with other \sts techiniques such as RL, agreement reranking and diverse decoding \citep{hassan2018achieving,ranzato2015sequence,li2016simple,baheti2018generating},
the proposed model is able to reach a  BLEU score of 32.3 for WMT 2017 Chinese-English translation, with a significant performance boost of +3.2 over existing SOTA results.

\section{Related Work}
Large scale pretraining has proved to be of significant importance in NLP, from word vectors 
such as word2vec/Glove in the early days 
\citep{pennington2014glove,mikolov2013distributed}
to recent language model pretraining such as BERT \citep{devlin2018bert} and Elmo \citep{peters2018deep}. 
Pretraining has not achieved comparable success in MT, mostly due to the difficulty of getting
a large
 parallel corpus. Most NMT pretraining work focuses on unsupervised NMT \citep{lample2019cross} or pretraining using monolingual data \citep{song2019mass}.  
Other relevant works for \sts pretraining include 
using an auto-encoder to pretrain the encoder-decoder network \citep{dai2015semi,ramachandran2016unsupervised} and
transfer learning from rich-source languages to low-source languages \citep{zoph2016transfer,firat2016zero}.

Our work is related to a couple of previous work in domain adaptation for MT. In 
the context of phrase-based MT,  \cite{hildebrand2005adaptation}
select sentences similar to the topic of the test set to construct a new training corpus, which avoids topic discrepancies between the training and the test datasets.
\cite{xiao2012topic} 
 estimate word translation probabilities conditioned on
topics, and then adapt lexical weights of phrases by these topic-specific probabilities.
In the context of NMT, \cite{chen2016guided}
provide neural networks with topic information (which are human-labeled product categories) on
the decoder side; 
\cite{zhang2016topic} first run topic models \citep{blei2003latent} on the training data for both sources and targets, and add topic representations to the encoder-decoder model.

Mixture models have been widely used in MT. 
In the context of phrase-based MT,
 \cite{foster2007mixture} propose a three-step pipelined strategy: they
 first split the training corpus into different sub-corpora  according to some predefined  criterion,
  then train different MT models on different sub-corpora, and in the end combine different models for translation. 
    \cite{foster2007mixture}'s work was later extended to address various issues \citep{niehues2010domain,koehn2007experiments,eidelman2012topic}, such as how to  split the training corpus \citep{axelrod2011domain} and
how to combine different models  \citep{civera2007domain,sennrich2012perplexity,foster2010discriminative}.
In NMT, mixture models \citep{shen2019mixture,he2018sequence} 
are inspired by deep latent variable generation models \citep{kingma2013auto,kim2018tutorial,bowman2015generating}.
\cite{zhang2016variational} 
 augment NMT systems with a single Gaussian latent
variable, and this work was further extended by \cite{schulz2018stochastic} in
which each target word is associated with a  latent Gaussian variable. 
\cite{he2018sequence} propose to use a soft mixture model to generate diverse translations. 
In addition, \cite{shen2019mixture} comprehensively evaluate different  design choices of mixture models such as parameterization and prior distribution. 

Due to the gigantic size of the dataset,  we have to split it into subsets during training. Therefore, our paper is also relevant to a wide range of previous work on 
distributed training for deep neural networks \citep{dean2012large,yadan2013multi,li2014scaling,krizhevsky2014one,das2016distributed,smith2017don}.

 \section{Data Setup}
The most commonly used Chinese-to-English (Zh-En) translation dataset is WMT'17. 
The dataset consists of 332K sentence pairs from the News Commentary corpus, 15.8M sentence pairs
from the UN Parallel Corpus, and 9M sentence pairs from the CWMT Corpus. 
We followed the pre-processing criterion in \cite{hassan2018achieving}, resulting in about 20M training pairs. 
Newsdev2017 is used as the development set and Newstest2017 as the test set. 

In addition to WMT2017,
we collected a Chinese-English parallel dataset that 
significantly extends those used
by  previous work. The raw dataset we collected consists of roughly 50 billion sentence pairs in total. The data comes from 
 diverse
sources such as web pages ($\sim$2 billion), digitized books ($\sim$1 billion) and private purchase from translation agencies  ($\sim$46 billion). 

The dataset is extremely noisy in two ways: 
(1) a significant proportion of files are in PDF format.
 (2)  text is aligned at the document level rather than the sentence level. The size of large documents can be up to thousands of pages, with figures, tables and charts annoyingly inserted.   
For the first part, we developed a  PDF document parsing system to decode PDF files. We are not diving into the details here since  this part are out of the scope of the current paper. Succinctly, a lexical analyzer is first used to decompose PDF files into the basic lexical tokens according to the PostScript syntax. Next, we build a parser to analyze the abstract syntax tree (AST), and then decode the data into figures, tables, and text. 
For the second part, our goal is to transform doc-level alignment to sentence-level alignment. 
We use a hierarchical pipeline which consists of two stages: (1) aligning paragraphs and (2) aligning sentences within each aligned paragraph. 
We adopt 
many of the techniques in \cite{uszkoreit2010large}:
Paragraphs/sentences are discarded  if both sides are identical or a language detector find them to be in the wrong language.
We
use a standard dynamic
programming approach to implement the sentence/paragraph alignment algorithm, which takes 
sentence length and translation probability as features. 
Pairs are discarded if pairing scores are less than a certain threshold. We encourage the readers to refer to \cite{uszkoreit2010large} for details. 
After post-processing, we are left with 41 billion sentence pairs. We randomly select 1M instances as the development set. 

\section{Models and Architectures}

\subsection{Pipelines: Pretraining and Fine-tuning}
Translation quality 
 is very vulnerable to out-of-domain data and noisy data \citep{chen2016guided,niehues2010domain,koehn2007experiments,eidelman2012topic}.
Since dataset we use comes from a different domain from WMT2017, 
it would be less favorable if we directly apply a trained model on this large but noisy dataset  to the WMT17 test set. 
One option to handle this issue is to do data selection and filtering before training \citep{belinkov2017synthetic,hassan2018achieving}. 
\cite{hassan2018achieving} proposes to first learn sentence representations from the provided training data in WMT2017 (target domain), and then reject training instances if their  
 similarity with the target sentences is below a prespecified
threshold. We did not choose this method for two reasons: 
(1) 
\cite{hassan2018achieving}  select different training instances for different target domains. 
 This means every time we encounter a new domain, we have to retrain the model; (2) the value of data filtering threshold is crucial but hard to decide:
it is unrealistic to tune its value since each threshold value corresponds to a different filtered training set, on which a brand new model has to be trained. 

Inspired by large-scale pretraining strategies such as BERT \citep{devlin2018bert} and Elmo \citep{peters2018deep}, we used a pipelined approach: we first pretrain an NMT model on the massive dataset, and then fine-tune the model 
on the training set of 
 the target domain. This strategy naturally addresses the aforementioned two caveats of data pre-filtering approach: the pretrained model can be easily adapted to a set of training data of an arbitrary domain, and it is no longer needed to find the optimal data selection threshold. 
Moreover, since the model will be fine-tuned at later stage, it is more immune to noise in the data at the first stage. 
 We combine the WMT 20M data with our new 40B data to do the pretraining, and then fine-tune the model on the WMT 20M data.

\subsection{Tradeoff between full-dataset optimality and convergence speed}
There is a tradeoff between the optimality on the full dataset and the convergence speed. At one end of the spectrum, running a single model on the full dataset may achieve full-dataset optimality. However, this strategy suffers from prohibitively long training time.\footnote{With a single V100 GPU, 
a single update on a single batch with about 2-5K tokens takes about 2-6 seconds.
Even with 512 parallel GPUs, it takes months for an epoch to finish.}
At the other end of the spectrum, splitting the full dataset into smaller subsets and running independent models on different datasets until convergence as in \cite{foster2007mixture} saves a lot of training time. But each individual model only finds the optimal solution for its own subset dataset, so their combination is not guaranteed to be the
  optimal model for the full dataset. 
We harvest the best of both worlds by model communication and data communication (see later sections for details).

\subsection{Model Details}
We use the Transformer architecture  \citep{vaswani2017attention}  as a backbone.
The
encoder and the decoder both have 6 blocks. 
The number of attention heads, embedding dimension and inner-layer dimension are set to 16, 1,024 and 2,048. We use the same transformer structure for all experiments. 
All experiments were run using 
512 Nvidia V100 GPUs with mini-batches
of approximately  1M tokens.  
Models are optimized with 
Adam \citep{kingma2014adam} in which $\beta_1$ is set to 0.9, $\beta_2$ is set to 0.98 and $\epsilon$ is set to $10^{-8}$. 
For the Chinese language, instead of using segmented words or byte-pair encoding (BPE) \citep{sennrich2015neural}, we use characters as the basic units and maintain a character vocabulary size of 10,000.  
We will get back to this in the ablation study section. 
On the English side,  we segment text into subword symbols using BPE \citep{sennrich2015neural} and maintain a vocabulary size of 40K. 

\subsubsection{Pretraining}
We explore the following different strategies for model pretraining. 
\paragraph{Single-Model} We use a single transformer model to fit all the training data. 
We use 
512 Nvidia V100 GPUs with mini-batches
of approximately  1M tokens. 
Models
are saved every 0.1 epoch. 
Upon the submission of this paper, training has lasted for three months, 2 epochs in total, and
perplexity on the development set is still dropping. 

\paragraph{Uniform-data-split} The disadvantages of {\it single-model}  are obvious: 
(1) It is prohibitively slow and (2) it is unclear whether a single-model is powerful enough to model the full training dataset. 
We thus followed 
\cite{foster2007mixture} to build mixture models, 
in which we first split 
the full dataset into a few subsets, and then train independent model components on different subsets. Here ``component" refers to the individual transformer in the mixture model setup. Using multiple components naturally increase  the model's capacity and expressivity.

We  randomly split the 40B training set into $K = 10$ subsets, denoted by $D_1, D_2, ..., D_K$. 
At training, different transformers are independently trained on different subsets
using parallel GPUs  
until convergence.
At test time, 
the probability of $p(y|x)$ can be written as follows:
\begin{equation}
p(y|x) = \sum_{z} p(y|z,x) p(z|x)
\end{equation}
where $z$ can be simply thought as the index of different subsets. 
$p(y|z,x)$ is characterized by the \sts component trained on   $D_z, z\in [1, K]$. 
We assume that $p(z|x)=1/K$ is  uniform.
The generation of target $y$ is thus the ensemble of the $K$ models.

\begin{figure*}
\includegraphics[width=5.5in]{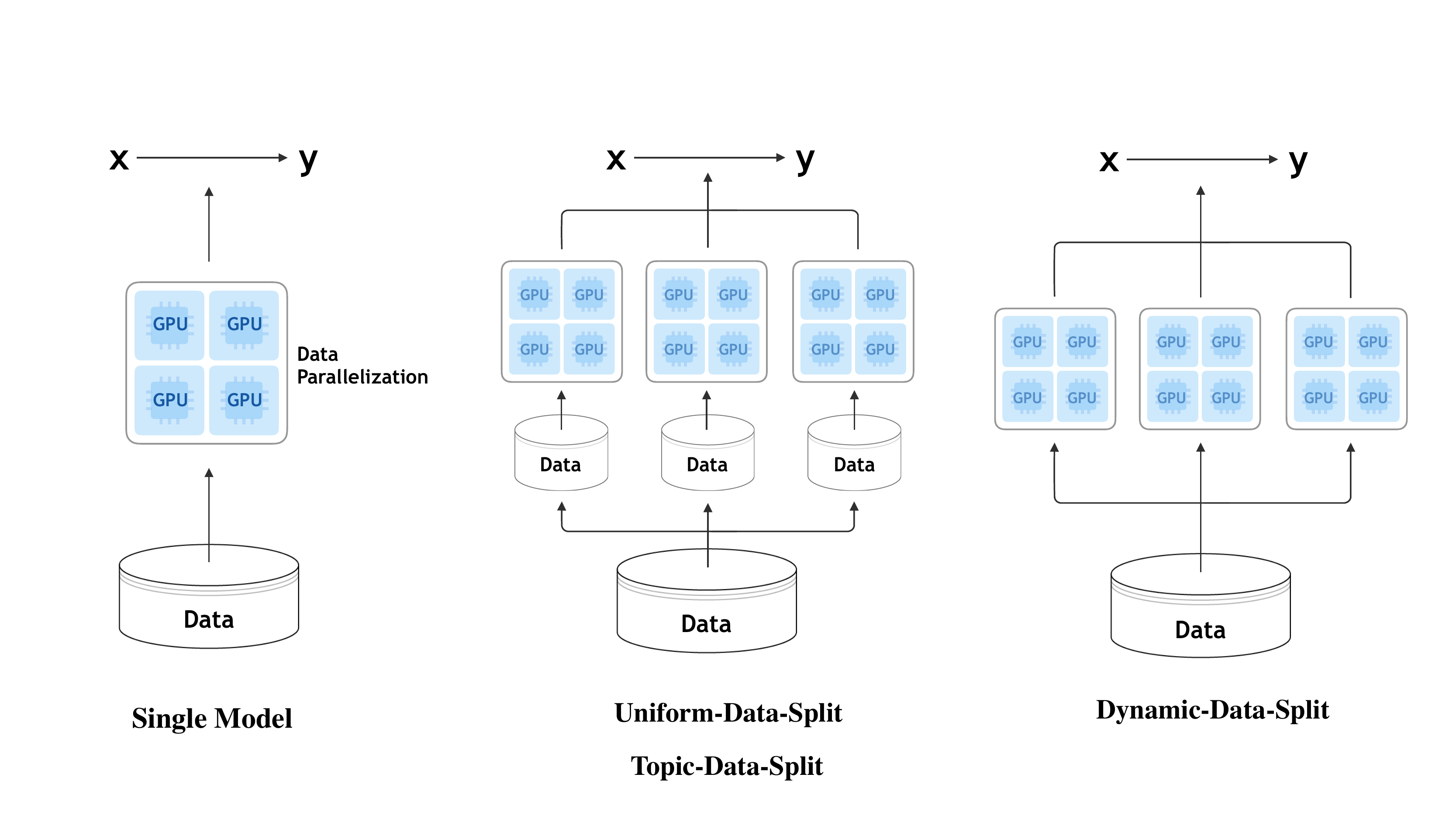}
\centering
\caption{Illustrations of different strategies for training.}
\label{overview}
\end{figure*}

\paragraph{Topic-data-split} 
The issue with {\it uniform-data-split} is that  subsets are randomly generated.
It would be more desirable if each $D_z$ represents a specific domain and  model components trained on different $D_z$  
are separate and have their own specialization and domain expertise. 
Domain-specific $D_z$ 
comes with the advantages of having
 fewer vocabularies, fewer low-frequency words and more 
sentences with similar topics and language expression patterns.  
We thus propose to split the full dataset in a more 
 elegant way using topic models \citep{blei2003latent}. 
Topic models are widely used in data split and selection in phrase-based MT \citep{hildebrand2005adaptation,zhao2006bitam,zhao2008hm}.
One tricky issue here is that each sentence pair consists of two different languages and we need 
to extract bilingual topics. 
To handle this issue, 
\cite{zhao2008hm} proposed
a  generative model, in which each source sentence is first sampled based on its topic, as in standard LDA, and then for each position in the source sentence, a target word is sampled based on a topic-specific translation lexicon. 
Variational EM is used for inference.  We refer the readers to \cite{zhao2008hm} for details. 
We followed \cite{zhao2008hm} and mined the topics distribution from the bilingual corpus. 
Each sentence pair is assigned to the subset that is most likely (has the highest probability) to contain the pair.

When data split is done, different \sts components are independently trained on different $D_z$. 
At test time, we use 
$p(y|x) = \sum_{z} p(y|z,x) p(z|x)$ for inference. 
$p(y|z,x)$ is characterized by the \sts component trained on  $D_z$.
Unlike {\it uniform-data-split}, $p(z|x)$ is not uniform, but predicted by a $K$-class classification model using BiLSTMs.
The classification model 
  first maps $x$ to a vector representation, which is then fed to a $K$-class softmax function. 

\paragraph{Dynamic-data-split} 
For 
{\it topic-data-split} and {\it uniform-data-split} strategies, 
subsets are generated beforehand and then fixed during the training. 
 We are thus unable to dynamically adjust the data during training. 
This means that if a training pair is assigned to a wrong subset, 
there is no way we can correct it and it will negatively affect the training process permanently. 
Inspired by mixture models for NMT \citep{shen2019mixture,he2018sequence},
we  propose to dynamically assign training instances to different model components, and update different components 
according to the examples assigned. 
\begin{equation}
\nabla_{\theta} \log\sum_z p(y|z,x)p(z|x) = \mathbb{E} _{p(z|x,y)} \nabla_{\theta}\log p(y,z|x)
\end{equation}
 We choose to use hard-EM instead of vanilla-EM due to the concern about the training speed: 
for vanilla-EM,
all components need to 
 run a forward-backward step on each training instance.\footnote{For different values of $z_i$, each $p(y_i|z_i, x_i)$ will be updated by gradients $\bigtriangledown_{\theta}\log p(y_i,z_i| x_i)$  weighted by the corresponding responsibility $p(z_i|x_i)$.} 
This is computationally intensive and we would like to avoid the slow convergence issue. 
Using hard-EM, models are trained  by iteratively applying the following two steps:

{\bf E-step}: For a given sentence pair $(x_i, y_i)$, estimate  $z_i$ by:
$z_i = \text{argmax}_z \log p(y_i|z_i, x_i)$,

{\bf M-step}: update parameters  with gradients $\bigtriangledown_{\theta} \log p(y_i, z_i| x_i)$.

We need to extend the single-instance EM step above to batched computation. 
There are two issues that requires special design:
 (1) the  notorious ``richer gets richer" issue \citep{eigen2013learning,shazeer2017outrageously,shen2019mixture}, i.e., once one  component is slightly better than the others, it will always be selected  
and other components will never get trained. 
Then the latent variable $z$  becomes useless and 
 the entire system degenerates to the vanilla \sts model that models $p(y|x)$; (2) 
 to accelerate training, we need all components to get updated all the time. 
Recall that each component have hundreds of parallel GPUs for computing. If some components do not get enough data, which is possible in dynamic-data-split, it would be a waste of computational resources.

We  propose the {\bf batched-E-step} strategy to deal with these two issues: suppose that the batch size for the mixture model component is $B$ (the value of which is approximately 1M). Since we have $K$ components, we feed $K\times B$ training instances to the model for each E-step. 
We ensure that each component is {\bf guaranteed} to be assigned $B$ instances 
in the E-step in order
 to get sufficiently updated in the subsequent M step. 
The batch-level assignment $z$ are computed as follows:
\begin{equation}
\begin{aligned}
& \max_{z_1, z_2, ..., z_{BK} } \sum_{i=1}^{i=BK} \log p(y_i|z_i, x_i) \\
&\text{s.t.}\sum_{i=1}^{BK}\mathbb{I} (z_i = z) = B ~~\text{for~} z = 1, 2, ..., K
\end{aligned}
\label{ILP}
\end{equation}
Eq. \ref{ILP} is an integer linear programming (ILP) problem. ILP is NP-hard  and is solved  using 
Hill Climbing \citep{russell2016artificial}, a heuristic method to find the optimal solution. 
The proposed strategy naturally
simultaneously 
 avoid the aforementioned ``richer gets richer" issue and 
the potential waste of computational resources. 
It is worth noting that the proposed {\bf batched-E-step} is not specific to our  scenario, but a general solution to the long-standing degeneracy issue of neural mixture models for text generation \citep{shazeer2017outrageously,shen2019mixture}. 

\subsubsection{Fine-Tuning}
At the model fine-tuning stage, we maintain the structure of the original pretrained model and fine-tune the model on the 20M WMT17 dataset. 
The number of iterations is treated as a hyper-parameter, which is tuned on the development set of WMT17. 

For {\it single-model}, we maintain the transformer structure and run additional iterations. 
For {\it uniform-data-split} and {\it topic-data-split}, 
we fine-tune 
each  component on the WMT17 dataset. 
At test time, constituent components are combined for decoding. 
Translations from 
{\it uniform-data-split} and {\it topic-data-split} are   generated by the ensemble of $K$ models. 

For {\it dynamic-data-split}, at the fine-tuning stage we run mixture models on the WMT17 dataset with minor adjustments: 
we replace the hard-EM and the batched-E-step
with  vanilla soft-EM, in which  all components  get updated with each training instance.
We do this because:
(1) The WMT17 dataset is significantly smaller so the computational cost is no longer a concern; (2) Mixture models have already been sufficiently trained during the pretraining, so we are less concerned about the ``richer gets richer" issue. Instead, it is even desirable that some  components get more fine-tuning if they are more relevant to the target domain. 

\begin{table}
\center
\begin{tabular}{llc}
Training Data & Setting  & Performance \\\hline
20M & Transformer \citep{hassan2018achieving}  & 24.4 \\
20M & Sogou \citep{wang2017sogou} & 26.4\\
20M & Microsoft  \citep{hassan2018achieving} & 27.4 \\
20M & Teacher Forcing \citep{he2019hard} & 29.1 \\
20M+100M&Microsoft  \citep{hassan2018achieving} &28.4 \\\hline
20M+40B (pretrain only) & single-model (1 epoch) & 22.1 \\
20M+40B (pretrain only) & single-model (2 epoch) & 24.7 \\
20M+40B (pretrain only) (ensemble) & Uniform Data Split & 27.2  \\
20M+40B (pretrain only) (ensemble)& Topic Data Split&27.7 \\
20M+40B (pretrain only) (ensemble)& Dynamic Data Split&28.4  \\\hline
20M+40B (pretrain+finetune)& single-model (1 epoch) & 27.4 \\
20M+40B (pretrain+finetune)& single-model (2 epoch) & 28.7 \\
20M+40B (pretrain+finetune) (ensemble)& Uniform Data Split & 31.1  \\
20M+40B (pretrain+finetune) (ensemble)& Topic Data Split&31.5 \\
\textbf{20M+40B (pretrain+finetune) (ensemble)}& \textbf{Dynamic Data Split} & \textbf{32.0}  \\\hline
\end{tabular}
\caption{Main results of different models and different settings  on the WMT 2017 Chinese-English test set.}
\label{result}
\end{table}

\section{Experimental Results}
Following \cite{hassan2018achieving,he2019hard}, we use 
 sacreBLEU\footnote{\url{https://github.com/awslabs/sockeye/tree/master/contrib/sacrebleu}} for evaluation. 
Results are reported in Table \ref{result}.

\subsection{Results}
\paragraph{Baselines} We copied baseline results from the Sogou best WMT17 system \citep{wang2017sogou}, the microsoft system \citep{hassan2018achieving}, and the current SOTA result using teaching forcing  \citep{he2019hard}.

\paragraph{Pretrain Only} 
We first take a look at results of the pretrain-only setting, in which we directly apply the pretrained models to the test set (it is worth noting that the pretrained training data contains the WMT17 training set). 
 For the {\it single-model}, due to the prohibitively long training time, we are only able to finish two epochs upon the submission of this paper, which has been running for 3 months. 
Though the model has not fully converged, 
its BLEU score (24.7) is slightly higher than the
performance of the same model trained on only the WMT17 dataset (24.4). 

Mixture models with different dataset split strategies outperform  {\it single-model} by a huge margin. This is due to the reasons that 
 (1) The {\it single-model}  has not fully converged yet; (2) 
 Model capacities of mixture models are significantly larger (scaled by $K$) than the {\it single-model}; and
   most importantly (3)  {\it uniform-data-split}  is actually an ensemble of multiple models and the comparison is not completely fair.\footnote{A completely fair comparison would be to use an ensemble of 10 {\it single-model}, each of which is trained on the 40B dataset. But this is very computationally prohibitive for us.}
By comparing {\it uniform-data-split}, {\it topic-data-split} and {\it dynamic-data-split},
we can  see that the way the full dataset is split has a significant effect on the performance.
 {\it Topic-data-split} significantly outperforms {\it  uniform-data-split}.
 This is because
 subsets in   {\it topic-data-split}  
 contain 
 fewer low-frequency words and more sentences with similar language expression patterns.
 The {\it dynamic-data-split} strategy dynamically adjusts the training data, which  lead to more coherent subsets, and thus  better performances. 
 \paragraph{Pretrain+Finetune} displays similar  patterns as {\it Pretrain-Only}: the {\it single-model} achieves a BLEU score of 29.7, already outperforming the current best system. This demonstrates the great power of large-scale pretraining. 
The gap between {\it single-model-1-epoch}, {\it single-model-2-epoch} and {\it uniform-data-split} is narrowed in the {\it Pretrain+Finetune} setting than in {\it Pretrain-Only}.  
The best setting, 
 {\it dynamic-data-split},  achieves a BLEU score of 32.0. 

\begin{figure*}[!ht]
\includegraphics[width=5.5in]{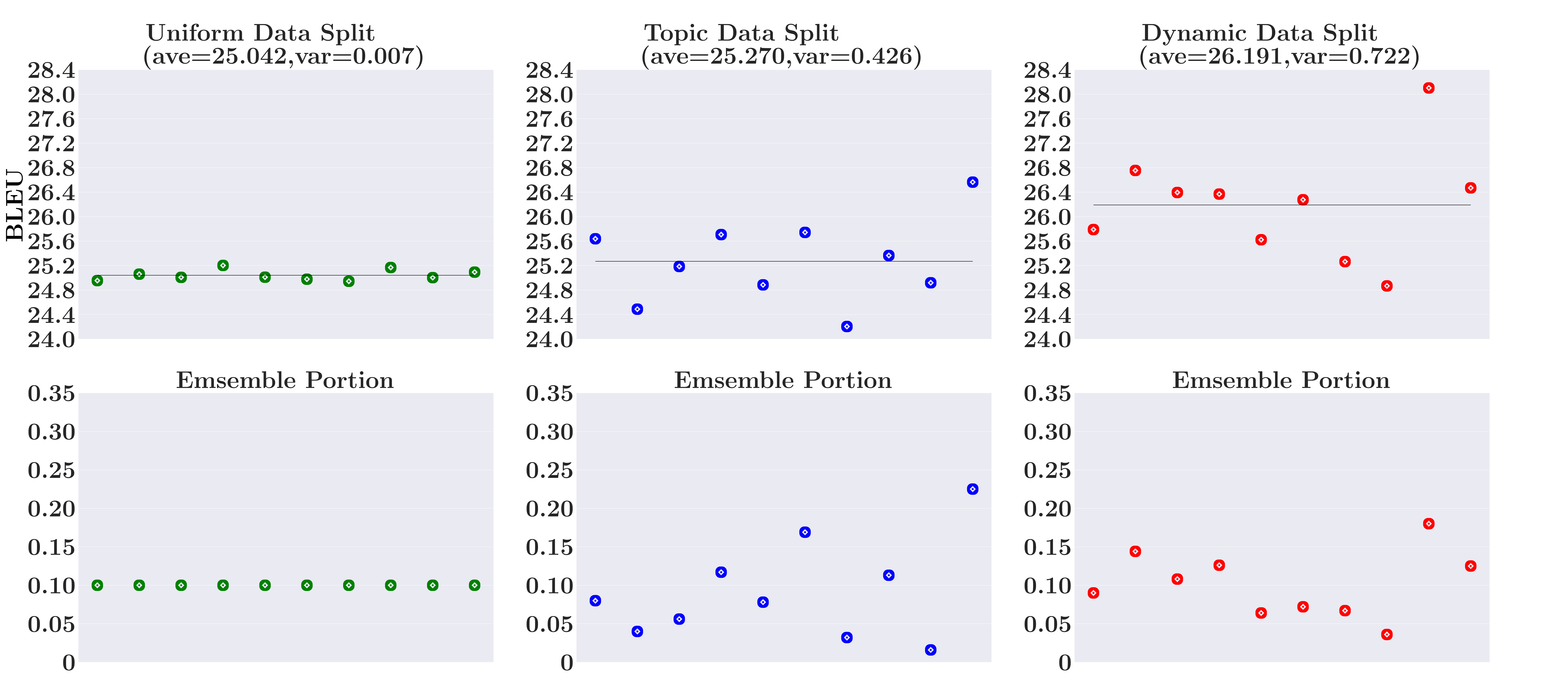}
\centering
\caption{Performances of each of the $K$ components in different data split strategies.}
\label{component}
\end{figure*}

\section{Ablation studies and analyses} 
\paragraph{Data Split Strategy}
In previous phrase-based MT work, both
how the full set is split \citep{axelrod2011domain} and how the mixture models are combined \citep{civera2007domain,sennrich2012perplexity,foster2010discriminative}
 have a significant impact on the final performance.
This is confirmed in our study. 
The first row of Figure \ref{component} corresponds to 
 the BLEU scores achieved by each of the $K$ components with the three data-split strategies (pretrain-only). 
As can be seen,
performances for different components in {\it uniform-data-split} are very similar. This is expected since the dataset is randomly split and all subsets come from the same distribution. The Variances 
in {\it topic-data-split} and {\it dynamic-data-split} are much larger.
This is because
subsets in  {\it topic-data-split} and {\it dynamic-data-split} have more centralized topic/domain distribution. Models trained on more 
specialized subsets perform much better than models trained on less specialized ones. 
The second row of Figure \ref{component} corresponds to the average weight of components in different mixture models. 
For  {\it uniform-data-split}, weight for each component is identical. For
{\it topic-data-split} and {\it dynamic-data-split}, we can find a high  correlation between weight and performance for each mixture component. This further explains the superiority of the two models. 

\paragraph{Model Size and Data Size}
It would be interesting to compare the performances of the following cases: (1) a large model trained on the full dataset (40B) but has not fully converged, i.e., {\it large-model-large-data}, (2) 
a converged large model on a subset of the dataset (4 Billion), i.e., {\it large-model-small-data} and (3) a model of a smaller size\footnote{Number of attention heads, embedding dimension and inner-layer dimension are set to 8, 512 and 512.}, which runs much faster on the full set (40B), i.e., {\it small-model-large-data}. 
Table \ref{size} represents the results. 
As can be seen, for the pretrain-only and the pretrain+finetune setting,  {\it small-model-large-data} performs the worst. When comparing Table \ref{size} with Table \ref{result},   
 {\it small-model-large-data} even underperforms a larger transformer trained only on the WMT17 dataset. 
 This demonstrates the 
 significant
 importance of model size and capacity. 
 Interestingly, 
   comparing  {\it large-model-large-data} with  {\it large-model-small-data},
the former performs a bit better on the pretrain-only setting, but performs a bit worse on the pretrain+fine-tune setting. 
Our explanation is as follows: for the pretrain-only setting,  {\it large-model-large-data} has not fully converged, and performs worse than the fully converged  {\it large-model-small-data} model. 
But in the pretrain+fine-tune setting, both models have fully converged on the WMT17 dataset. 
Since
 {\it large-model-large-data} was exposed to more data during the pretraining process, it has more generalization capability and achieves better performances during fine-tuning.

\begin{table}
\center
\begin{tabular}{llccc}\hline
Model & Dataset & Sufficiently Pretrained & Setting & BLEU \\\hline
Large & Large (40B) & No & Pretrain-Only &  24.7 \\
Large & Small (4B) & Yes & Pretrain-Only &  25.1 \\
Small & Large (40B) & Yes & Pretrain-Only &  23.2 \\\hline
Large & Large (40B) & No & Pretrain+Finetune &  28.7 \\
Large & Small (4B) & Yes & Pretrain+Finetune &  28.4 \\
Small & Large (40B) & Yes & Pretrain+Finetune &  26.8 \\\hline
\end{tabular}
\caption{Performances of non-ensemble models with different sizes and pre-trained on various amount of  data.}
\label{size}
\end{table}

\paragraph{Character, Word or BPE}
In large-scale neural network learning,
the necessity of  Chinese word segmentation (CWS) is still under huge debate \citep{meng2019word}.
The answer is not immediately clear because,
on the one hand, the data sparsity issue with word-based models 
 is less severe when massive training data is available, so word-based models might be useful. One the other hand, ``word" is a human-defined linguistic concept, characterized by labeled CWS datasets \citep{xia2000part,yu2001processing}. It is widely accepted that 
large-scale text learning can automatically learn and encode linguistic structures \citep{hinoshita2011emergence,williams2018latent}, which means CWS might be less useful with large training data. 
Using the 4B subset on which the best performance is achieved in the {\it topic-data-split} setting, we examine the performances of
 the word-based model (vocabulary set to 50K), the subword BPE model (vocabulary set to 50K) and the char-based model (vocabulary set to 10K). The three models only differ in the encoding stage. 
Results are shown in Table \ref{ab}. 
Combined with fine-tuning, the three models achieve BLEU scores of 
29.4, 29.8 and 30.1, respectively. This confirms the finding in \cite{meng2019word} that CWS is not necessary in NMT when large-scale training data is used.

\paragraph{Examination of Existing NMT Techniques}
It would be interesting to see  which existing widely-used
NMT 
techniques are still effective. 
Since our bilingual dataset is extremely larger than previous datasets, existing data augmentation strategies such as  
{\it monolingual  language model fusion} \citep{sennrich2015improving} or {\it back translation} \citep{hassan2018achieving,edunov2018understanding} 
are expected to be no longer effective. We use 100M monolingual data to verify this hypothesis. 
Other NMT techniques we examined include the following: 

Agreement Reranking: \cite{hassan2018achieving,li2015diversity}  reranks the N-best list using models that generate sources and targets from different directions, i.e., 
 S2T-L2R (target sentence
is generated from left to right), S2T-R2L, T2S-L2R and T2SR2L.
Due to the computational cost, we  only pretain S2T-R2L, T2S-L2R and T2SR2L  on the 4B dataset and then fine-tune them on  
WMT2017. 

Reinforce Learning \citep{ranzato2015sequence,wu2016google}: refining the \sts objective with the RL objective to  directly optimize the translation BLEU scores. We apply RL at the fine-tuning stage. 

Diverse Decoding \citep{li2017modeling,vijayakumar2018diverse}: using a diversity-promoting beam search, in which inter-sibling scores are penalized in order to generate more diverse N-best list.  

Results are shown in Table \ref{ab}. As can be seen,  monolingual  ML fusion and back-translation actually harm the performance. This is expected since the 
monolingual 
dataset is significantly smaller than the pretraining bilingual datasets. 
Agreement reranking introduces a +0.20 BLEU boost, confirming the importance of the order by which sequences are generated. 
The improvement from diverse decoding is also tiny. We think that this might because the model is already good enough for using beam search to find the global optimal. 
Another significant improvement comes from the RL strategy, leading to a roughly +0.2 BLEU boost.
The combination of RL, agreement-ranking and diverse decoding pushes the performance up to 32.3.  

\begin{table}
\center
\begin{tabular}{lc}\hline
\multicolumn{2}{c}{Word, BPE and Char}\\\hline
Word & 29.4 \\
BPE & 29.8 \\
Char & 30.1 \\\hline
\end{tabular}
~~~~~~~
\begin{tabular}{lc}\hline
\multicolumn{2}{c}{Different NMT Techniques}\\\hline
Currently best system & 32.02 \\
Monolingual  LM Fusion& 30.16 \\
Back Translation & 30.87 \\
Agreement  Reranking & 32.22 \\
RL & 32.20\\
Diverse Decoding & 32.11 \\
Agreement  Reranking + RL+ Diverse &32.29 \\\hline
\end{tabular}
\caption{(a) Results for word-based, BPE-based and char-based models. (b) Results for different existing NMT techniques. }
\label{ab}
\end{table}

\end{CJK*}

\section{Conclusion}
In this paper, we empirically study training NMT systems with 40B training instances, the size of which is by orders of magnitude larger than the largest dataset to date.
We provide practical solutions to handle the tradeoff between full-dataset level optimality and fast training speed, and demonstrate that large-scale pretraining significantly improves NMT performances. We are able to achieve a BLEU score of  32.3 on WMT17 Chinese-English dataset,
with a significant performance boost of +3.2 over existing SOTA results.

\bibliography{iclr2019_conference}

\begin{thebibliography}{72}
\providecommand{\natexlab}[1]{#1}
\providecommand{\url}[1]{\texttt{#1}}
\expandafter\ifx\csname urlstyle\endcsname\relax
  \providecommand{\doi}[1]{doi: #1}\else
  \providecommand{\doi}{doi: \begingroup \urlstyle{rm}\Url}\fi

\bibitem[Aharoni et~al.(2019)Aharoni, Johnson, and Firat]{aharoni2019massively}
Roee Aharoni, Melvin Johnson, and Orhan Firat.
\newblock Massively multilingual neural machine translation.
\newblock \emph{arXiv preprint arXiv:1903.00089}, 2019.

\bibitem[Axelrod et~al.(2011)Axelrod, He, and Gao]{axelrod2011domain}
Amittai Axelrod, Xiaodong He, and Jianfeng Gao.
\newblock Domain adaptation via pseudo in-domain data selection.
\newblock In \emph{Proceedings of the conference on empirical methods in
  natural language processing}, pp.\  355--362. Association for Computational
  Linguistics, 2011.

\bibitem[Bahdanau et~al.(2014)Bahdanau, Cho, and Bengio]{bahdanau2014neural}
Dzmitry Bahdanau, Kyunghyun Cho, and Yoshua Bengio.
\newblock Neural machine translation by jointly learning to align and
  translate.
\newblock \emph{arXiv preprint arXiv:1409.0473}, 2014.

\bibitem[Baheti et~al.(2018)Baheti, Ritter, Li, and
  Dolan]{baheti2018generating}
Ashutosh Baheti, Alan Ritter, Jiwei Li, and Bill Dolan.
\newblock Generating more interesting responses in neural conversation models
  with distributional constraints.
\newblock \emph{arXiv preprint arXiv:1809.01215}, 2018.

\bibitem[Belinkov \& Bisk(2017)Belinkov and Bisk]{belinkov2017synthetic}
Yonatan Belinkov and Yonatan Bisk.
\newblock Synthetic and natural noise both break neural machine translation.
\newblock \emph{arXiv preprint arXiv:1711.02173}, 2017.

\bibitem[Blei et~al.(2003)Blei, Ng, and Jordan]{blei2003latent}
David~M Blei, Andrew~Y Ng, and Michael~I Jordan.
\newblock Latent dirichlet allocation.
\newblock \emph{Journal of machine Learning research}, 3\penalty0
  (Jan):\penalty0 993--1022, 2003.

\bibitem[Bowman et~al.(2015)Bowman, Vilnis, Vinyals, Dai, Jozefowicz, and
  Bengio]{bowman2015generating}
Samuel~R Bowman, Luke Vilnis, Oriol Vinyals, Andrew~M Dai, Rafal Jozefowicz,
  and Samy Bengio.
\newblock Generating sentences from a continuous space.
\newblock \emph{arXiv preprint arXiv:1511.06349}, 2015.

\bibitem[Britz et~al.(2017)Britz, Goldie, Luong, and Le]{britz2017massive}
Denny Britz, Anna Goldie, Minh-Thang Luong, and Quoc Le.
\newblock Massive exploration of neural machine translation architectures.
\newblock \emph{arXiv preprint arXiv:1703.03906}, 2017.

\bibitem[Chen et~al.(2016)Chen, Matusov, Khadivi, and Peter]{chen2016guided}
Wenhu Chen, Evgeny Matusov, Shahram Khadivi, and Jan-Thorsten Peter.
\newblock Guided alignment training for topic-aware neural machine translation.
\newblock \emph{arXiv preprint arXiv:1607.01628}, 2016.

\bibitem[Civera \& Juan(2007)Civera and Juan]{civera2007domain}
Jorge Civera and Alfons Juan.
\newblock Domain adaptation in statistical machine translation with mixture
  modelling.
\newblock In \emph{Proceedings of the Second Workshop on Statistical Machine
  Translation}, pp.\  177--180. Association for Computational Linguistics,
  2007.

\bibitem[Dai \& Le(2015)Dai and Le]{dai2015semi}
Andrew~M Dai and Quoc~V Le.
\newblock Semi-supervised sequence learning.
\newblock In \emph{Advances in neural information processing systems}, pp.\
  3079--3087, 2015.

\bibitem[Das et~al.(2016)Das, Avancha, Mudigere, Vaidynathan, Sridharan,
  Kalamkar, Kaul, and Dubey]{das2016distributed}
Dipankar Das, Sasikanth Avancha, Dheevatsa Mudigere, Karthikeyan Vaidynathan,
  Srinivas Sridharan, Dhiraj Kalamkar, Bharat Kaul, and Pradeep Dubey.
\newblock Distributed deep learning using synchronous stochastic gradient
  descent.
\newblock \emph{arXiv preprint arXiv:1602.06709}, 2016.

\bibitem[Dean et~al.(2012)Dean, Corrado, Monga, Chen, Devin, Mao, Senior,
  Tucker, Yang, Le, et~al.]{dean2012large}
Jeffrey Dean, Greg Corrado, Rajat Monga, Kai Chen, Matthieu Devin, Mark Mao,
  Andrew Senior, Paul Tucker, Ke~Yang, Quoc~V Le, et~al.
\newblock Large scale distributed deep networks.
\newblock In \emph{Advances in neural information processing systems}, pp.\
  1223--1231, 2012.

\bibitem[Devlin et~al.(2018)Devlin, Chang, Lee, and Toutanova]{devlin2018bert}
Jacob Devlin, Ming-Wei Chang, Kenton Lee, and Kristina Toutanova.
\newblock Bert: Pre-training of deep bidirectional transformers for language
  understanding.
\newblock \emph{arXiv preprint arXiv:1810.04805}, 2018.

\bibitem[Edunov et~al.(2018)Edunov, Ott, Auli, and
  Grangier]{edunov2018understanding}
Sergey Edunov, Myle Ott, Michael Auli, and David Grangier.
\newblock Understanding back-translation at scale.
\newblock \emph{arXiv preprint arXiv:1808.09381}, 2018.

\bibitem[Eidelman et~al.(2012)Eidelman, Boyd-Graber, and
  Resnik]{eidelman2012topic}
Vladimir Eidelman, Jordan Boyd-Graber, and Philip Resnik.
\newblock Topic models for dynamic translation model adaptation.
\newblock In \emph{Proceedings of the 50th Annual Meeting of the Association
  for Computational Linguistics: Short Papers-Volume 2}, pp.\  115--119.
  Association for Computational Linguistics, 2012.

\bibitem[Eigen et~al.(2013)Eigen, Ranzato, and Sutskever]{eigen2013learning}
David Eigen, Marc'Aurelio Ranzato, and Ilya Sutskever.
\newblock Learning factored representations in a deep mixture of experts.
\newblock \emph{arXiv preprint arXiv:1312.4314}, 2013.

\bibitem[Firat et~al.(2016)Firat, Sankaran, Al-Onaizan, Vural, and
  Cho]{firat2016zero}
Orhan Firat, Baskaran Sankaran, Yaser Al-Onaizan, Fatos T~Yarman Vural, and
  Kyunghyun Cho.
\newblock Zero-resource translation with multi-lingual neural machine
  translation.
\newblock \emph{arXiv preprint arXiv:1606.04164}, 2016.

\bibitem[Foster \& Kuhn(2007)Foster and Kuhn]{foster2007mixture}
George Foster and Roland Kuhn.
\newblock Mixture-model adaptation for smt.
\newblock In \emph{Proceedings of the Second Workshop on Statistical Machine
  Translation}, pp.\  128--135. Association for Computational Linguistics,
  2007.

\bibitem[Foster et~al.(2010)Foster, Goutte, and Kuhn]{foster2010discriminative}
George Foster, Cyril Goutte, and Roland Kuhn.
\newblock Discriminative instance weighting for domain adaptation in
  statistical machine translation.
\newblock In \emph{Proceedings of the 2010 conference on empirical methods in
  natural language processing}, pp.\  451--459. Association for Computational
  Linguistics, 2010.

\bibitem[Gehring et~al.(2017)Gehring, Auli, Grangier, Yarats, and
  Dauphin]{gehring2017convolutional}
Jonas Gehring, Michael Auli, David Grangier, Denis Yarats, and Yann~N Dauphin.
\newblock Convolutional sequence to sequence learning.
\newblock In \emph{Proceedings of the 34th International Conference on Machine
  Learning-Volume 70}, pp.\  1243--1252. JMLR. org, 2017.

\bibitem[Hassan et~al.(2018)Hassan, Aue, Chen, Chowdhary, Clark, Federmann,
  Huang, Junczys-Dowmunt, Lewis, Li, et~al.]{hassan2018achieving}
Hany Hassan, Anthony Aue, Chang Chen, Vishal Chowdhary, Jonathan Clark,
  Christian Federmann, Xuedong Huang, Marcin Junczys-Dowmunt, William Lewis,
  Mu~Li, et~al.
\newblock Achieving human parity on automatic chinese to english news
  translation.
\newblock \emph{arXiv preprint arXiv:1803.05567}, 2018.

\bibitem[He et~al.(2019)He, Tan, and Qin]{he2019hard}
Tianyu He, Xu~Tan, and Tao Qin.
\newblock Hard but robust, easy but sensitive: How encoder and decoder perform
  in neural machine translation.
\newblock \emph{arXiv preprint arXiv:1908.06259}, 2019.

\bibitem[He et~al.(2018)He, Haffari, and Norouzi]{he2018sequence}
Xuanli He, Gholamreza Haffari, and Mohammad Norouzi.
\newblock Sequence to sequence mixture model for diverse machine translation.
\newblock \emph{arXiv preprint arXiv:1810.07391}, 2018.

\bibitem[Hildebrand et~al.(2005)Hildebrand, Eck, Vogel, and
  Waibel]{hildebrand2005adaptation}
Almut~Silja Hildebrand, Matthias Eck, Stephan Vogel, and Alex Waibel.
\newblock Adaptation of the translation model for statistical machine
  translation based on information retrieval.
\newblock In \emph{Proceedings of EAMT}, volume 2005, pp.\  133--142, 2005.

\bibitem[Hinoshita et~al.(2011)Hinoshita, Arie, Tani, Okuno, and
  Ogata]{hinoshita2011emergence}
Wataru Hinoshita, Hiroaki Arie, Jun Tani, Hiroshi~G Okuno, and Tetsuya Ogata.
\newblock Emergence of hierarchical structure mirroring linguistic composition
  in a recurrent neural network.
\newblock \emph{Neural Networks}, 24\penalty0 (4):\penalty0 311--320, 2011.

\bibitem[Johnson et~al.(2017)Johnson, Schuster, Le, Krikun, Wu, Chen, Thorat,
  Vi{\'e}gas, Wattenberg, Corrado, et~al.]{johnson2017google}
Melvin Johnson, Mike Schuster, Quoc~V Le, Maxim Krikun, Yonghui Wu, Zhifeng
  Chen, Nikhil Thorat, Fernanda Vi{\'e}gas, Martin Wattenberg, Greg Corrado,
  et~al.
\newblock Google’s multilingual neural machine translation system: Enabling
  zero-shot translation.
\newblock \emph{Transactions of the Association for Computational Linguistics},
  5:\penalty0 339--351, 2017.

\bibitem[Kim et~al.(2018)Kim, Wiseman, and Rush]{kim2018tutorial}
Yoon Kim, Sam Wiseman, and Alexander~M Rush.
\newblock A tutorial on deep latent variable models of natural language.
\newblock \emph{arXiv preprint arXiv:1812.06834}, 2018.

\bibitem[Kingma \& Ba(2014)Kingma and Ba]{kingma2014adam}
Diederik~P Kingma and Jimmy Ba.
\newblock Adam: A method for stochastic optimization.
\newblock \emph{arXiv preprint arXiv:1412.6980}, 2014.

\bibitem[Kingma \& Welling(2013)Kingma and Welling]{kingma2013auto}
Diederik~P Kingma and Max Welling.
\newblock Auto-encoding variational bayes.
\newblock \emph{arXiv preprint arXiv:1312.6114}, 2013.

\bibitem[Klein et~al.(2017)Klein, Kim, Deng, Senellart, and
  Rush]{klein2017opennmt}
Guillaume Klein, Yoon Kim, Yuntian Deng, Jean Senellart, and Alexander~M Rush.
\newblock Opennmt: Open-source toolkit for neural machine translation.
\newblock \emph{arXiv preprint arXiv:1701.02810}, 2017.

\bibitem[Koehn \& Schroeder(2007)Koehn and Schroeder]{koehn2007experiments}
Philipp Koehn and Josh Schroeder.
\newblock Experiments in domain adaptation for statistical machine translation.
\newblock In \emph{Proceedings of the second workshop on statistical machine
  translation}, pp.\  224--227, 2007.

\bibitem[Krizhevsky(2014)]{krizhevsky2014one}
Alex Krizhevsky.
\newblock One weird trick for parallelizing convolutional neural networks.
\newblock \emph{arXiv preprint arXiv:1404.5997}, 2014.

\bibitem[Lample \& Conneau(2019)Lample and Conneau]{lample2019cross}
Guillaume Lample and Alexis Conneau.
\newblock Cross-lingual language model pretraining.
\newblock \emph{arXiv preprint arXiv:1901.07291}, 2019.

\bibitem[Li et~al.(2015)Li, Galley, Brockett, Gao, and Dolan]{li2015diversity}
Jiwei Li, Michel Galley, Chris Brockett, Jianfeng Gao, and Bill Dolan.
\newblock A diversity-promoting objective function for neural conversation
  models.
\newblock \emph{arXiv preprint arXiv:1510.03055}, 2015.

\bibitem[Li et~al.(2016)Li, Monroe, and Jurafsky]{li2016simple}
Jiwei Li, Will Monroe, and Dan Jurafsky.
\newblock A simple, fast diverse decoding algorithm for neural generation.
\newblock \emph{arXiv preprint arXiv:1611.08562}, 2016.

\bibitem[Li et~al.(2017)Li, Xiong, Tu, Zhu, Zhang, and Zhou]{li2017modeling}
Junhui Li, Deyi Xiong, Zhaopeng Tu, Muhua Zhu, Min Zhang, and Guodong Zhou.
\newblock Modeling source syntax for neural machine translation.
\newblock \emph{arXiv preprint arXiv:1705.01020}, 2017.

\bibitem[Li et~al.(2014)Li, Andersen, Park, Smola, Ahmed, Josifovski, Long,
  Shekita, and Su]{li2014scaling}
Mu~Li, David~G Andersen, Jun~Woo Park, Alexander~J Smola, Amr Ahmed, Vanja
  Josifovski, James Long, Eugene~J Shekita, and Bor-Yiing Su.
\newblock Scaling distributed machine learning with the parameter server.
\newblock In \emph{11th $\{$USENIX$\}$ Symposium on Operating Systems Design
  and Implementation ($\{$OSDI$\}$ 14)}, pp.\  583--598, 2014.

\bibitem[Luong et~al.(2015)Luong, Pham, and Manning]{luong2015effective}
Minh-Thang Luong, Hieu Pham, and Christopher~D Manning.
\newblock Effective approaches to attention-based neural machine translation.
\newblock \emph{arXiv preprint arXiv:1508.04025}, 2015.

\bibitem[Meng et~al.(2019)Meng, Li, Sun, Han, Yuan, and Li]{meng2019word}
Yuxian Meng, Xiaoya Li, Xiaofei Sun, Qinghong Han, Arianna Yuan, and Jiwei Li.
\newblock Is word segmentation necessary for deep learning of chinese
  representations?
\newblock \emph{arXiv preprint arXiv:1905.05526}, 2019.

\bibitem[Mikolov et~al.(2013)Mikolov, Sutskever, Chen, Corrado, and
  Dean]{mikolov2013distributed}
Tomas Mikolov, Ilya Sutskever, Kai Chen, Greg~S Corrado, and Jeff Dean.
\newblock Distributed representations of words and phrases and their
  compositionality.
\newblock In \emph{Advances in neural information processing systems}, pp.\
  3111--3119, 2013.

\bibitem[Ng et~al.(2019)Ng, Yee, Baevski, Ott, Auli, and
  Edunov]{ng2019facebook}
Nathan Ng, Kyra Yee, Alexei Baevski, Myle Ott, Michael Auli, and Sergey Edunov.
\newblock Facebook fair's wmt19 news translation task submission.
\newblock \emph{arXiv preprint arXiv:1907.06616}, 2019.

\bibitem[Niehues \& Waibel(2010)Niehues and Waibel]{niehues2010domain}
Jan Niehues and Alex Waibel.
\newblock Domain adaptation in statistical machine translation using factored
  translation models.
\newblock In \emph{Proceedings of EAMT}, 2010.

\bibitem[Pennington et~al.(2014)Pennington, Socher, and
  Manning]{pennington2014glove}
Jeffrey Pennington, Richard Socher, and Christopher Manning.
\newblock Glove: Global vectors for word representation.
\newblock In \emph{Proceedings of the 2014 conference on empirical methods in
  natural language processing (EMNLP)}, pp.\  1532--1543, 2014.

\bibitem[Peters et~al.(2018)Peters, Neumann, Iyyer, Gardner, Clark, Lee, and
  Zettlemoyer]{peters2018deep}
Matthew~E Peters, Mark Neumann, Mohit Iyyer, Matt Gardner, Christopher Clark,
  Kenton Lee, and Luke Zettlemoyer.
\newblock Deep contextualized word representations.
\newblock \emph{arXiv preprint arXiv:1802.05365}, 2018.

\bibitem[Ramachandran et~al.(2016)Ramachandran, Liu, and
  Le]{ramachandran2016unsupervised}
Prajit Ramachandran, Peter~J Liu, and Quoc~V Le.
\newblock Unsupervised pretraining for sequence to sequence learning.
\newblock \emph{arXiv preprint arXiv:1611.02683}, 2016.

\bibitem[Ranzato et~al.(2015)Ranzato, Chopra, Auli, and
  Zaremba]{ranzato2015sequence}
Marc'Aurelio Ranzato, Sumit Chopra, Michael Auli, and Wojciech Zaremba.
\newblock Sequence level training with recurrent neural networks.
\newblock \emph{arXiv preprint arXiv:1511.06732}, 2015.

\bibitem[Russell \& Norvig(2016)Russell and Norvig]{russell2016artificial}
Stuart~J Russell and Peter Norvig.
\newblock \emph{Artificial intelligence: a modern approach}.
\newblock Malaysia; Pearson Education Limited,, 2016.

\bibitem[Schulz et~al.(2018)Schulz, Aziz, and Cohn]{schulz2018stochastic}
Philip Schulz, Wilker Aziz, and Trevor Cohn.
\newblock A stochastic decoder for neural machine translation.
\newblock \emph{arXiv preprint arXiv:1805.10844}, 2018.

\bibitem[Sennrich(2012)]{sennrich2012perplexity}
Rico Sennrich.
\newblock Perplexity minimization for translation model domain adaptation in
  statistical machine translation.
\newblock In \emph{Proceedings of the 13th Conference of the European Chapter
  of the Association for Computational Linguistics}, pp.\  539--549.
  Association for Computational Linguistics, 2012.

\bibitem[Sennrich et~al.(2015{\natexlab{a}})Sennrich, Haddow, and
  Birch]{sennrich2015improving}
Rico Sennrich, Barry Haddow, and Alexandra Birch.
\newblock Improving neural machine translation models with monolingual data.
\newblock \emph{arXiv preprint arXiv:1511.06709}, 2015{\natexlab{a}}.

\bibitem[Sennrich et~al.(2015{\natexlab{b}})Sennrich, Haddow, and
  Birch]{sennrich2015neural}
Rico Sennrich, Barry Haddow, and Alexandra Birch.
\newblock Neural machine translation of rare words with subword units.
\newblock \emph{arXiv preprint arXiv:1508.07909}, 2015{\natexlab{b}}.

\bibitem[Shazeer et~al.(2017)Shazeer, Mirhoseini, Maziarz, Davis, Le, Hinton,
  and Dean]{shazeer2017outrageously}
Noam Shazeer, Azalia Mirhoseini, Krzysztof Maziarz, Andy Davis, Quoc Le,
  Geoffrey Hinton, and Jeff Dean.
\newblock Outrageously large neural networks: The sparsely-gated
  mixture-of-experts layer.
\newblock \emph{arXiv preprint arXiv:1701.06538}, 2017.

\bibitem[Shen et~al.(2019)Shen, Ott, Auli, and Ranzato]{shen2019mixture}
Tianxiao Shen, Myle Ott, Michael Auli, and Marc'Aurelio Ranzato.
\newblock Mixture models for diverse machine translation: Tricks of the trade.
\newblock \emph{arXiv preprint arXiv:1902.07816}, 2019.

\bibitem[Smith et~al.(2017)Smith, Kindermans, Ying, and Le]{smith2017don}
Samuel~L Smith, Pieter-Jan Kindermans, Chris Ying, and Quoc~V Le.
\newblock Don't decay the learning rate, increase the batch size.
\newblock \emph{arXiv preprint arXiv:1711.00489}, 2017.

\bibitem[Song et~al.(2019)Song, Tan, Qin, Lu, and Liu]{song2019mass}
Kaitao Song, Xu~Tan, Tao Qin, Jianfeng Lu, and Tie-Yan Liu.
\newblock Mass: Masked sequence to sequence pre-training for language
  generation.
\newblock \emph{arXiv preprint arXiv:1905.02450}, 2019.

\bibitem[Sutskever et~al.(2014)Sutskever, Vinyals, and
  Le]{sutskever2014sequence}
Ilya Sutskever, Oriol Vinyals, and Quoc~V Le.
\newblock Sequence to sequence learning with neural networks.
\newblock In \emph{Advances in neural information processing systems}, pp.\
  3104--3112, 2014.

\bibitem[Uszkoreit et~al.(2010)Uszkoreit, Ponte, Popat, and
  Dubiner]{uszkoreit2010large}
Jakob Uszkoreit, Jay~M Ponte, Ashok~C Popat, and Moshe Dubiner.
\newblock Large scale parallel document mining for machine translation.
\newblock In \emph{Proceedings of the 23rd International Conference on
  Computational Linguistics}, pp.\  1101--1109. Association for Computational
  Linguistics, 2010.

\bibitem[Vaswani et~al.(2017)Vaswani, Shazeer, Parmar, Uszkoreit, Jones, Gomez,
  Kaiser, and Polosukhin]{vaswani2017attention}
Ashish Vaswani, Noam Shazeer, Niki Parmar, Jakob Uszkoreit, Llion Jones,
  Aidan~N Gomez, {\L}ukasz Kaiser, and Illia Polosukhin.
\newblock Attention is all you need.
\newblock In \emph{Advances in neural information processing systems}, pp.\
  5998--6008, 2017.

\bibitem[Vijayakumar et~al.(2018)Vijayakumar, Cogswell, Selvaraju, Sun, Lee,
  Crandall, and Batra]{vijayakumar2018diverse}
Ashwin~K Vijayakumar, Michael Cogswell, Ramprasaath~R Selvaraju, Qing Sun,
  Stefan Lee, David Crandall, and Dhruv Batra.
\newblock Diverse beam search for improved description of complex scenes.
\newblock In \emph{Thirty-Second AAAI Conference on Artificial Intelligence},
  2018.

\bibitem[Wang et~al.(2017)Wang, Cheng, Jiang, Yang, Chen, Li, Shi, Wang, and
  Yang]{wang2017sogou}
Yuguang Wang, Shanbo Cheng, Liyang Jiang, Jiajun Yang, Wei Chen, Muze Li, Lin
  Shi, Yanfeng Wang, and Hongtao Yang.
\newblock Sogou neural machine translation systems for wmt17.
\newblock In \emph{Proceedings of the Second Conference on Machine
  Translation}, pp.\  410--415, 2017.

\bibitem[Williams et~al.(2018)Williams, Drozdov*, and
  Bowman]{williams2018latent}
Adina Williams, Andrew Drozdov*, and Samuel~R Bowman.
\newblock Do latent tree learning models identify meaningful structure in
  sentences?
\newblock \emph{Transactions of the Association for Computational Linguistics},
  6:\penalty0 253--267, 2018.

\bibitem[Wu et~al.(2016)Wu, Schuster, Chen, Le, Norouzi, Macherey, Krikun, Cao,
  Gao, Macherey, et~al.]{wu2016google}
Yonghui Wu, Mike Schuster, Zhifeng Chen, Quoc~V Le, Mohammad Norouzi, Wolfgang
  Macherey, Maxim Krikun, Yuan Cao, Qin Gao, Klaus Macherey, et~al.
\newblock Google's neural machine translation system: Bridging the gap between
  human and machine translation.
\newblock \emph{arXiv preprint arXiv:1609.08144}, 2016.

\bibitem[Xia(2000)]{xia2000part}
Fei Xia.
\newblock The part-of-speech tagging guidelines for the penn chinese treebank
  (3.0).
\newblock \emph{IRCS Technical Reports Series}, pp.\ ~38, 2000.

\bibitem[Xiao et~al.(2012)Xiao, Xiong, Zhang, Liu, and Lin]{xiao2012topic}
Xinyan Xiao, Deyi Xiong, Min Zhang, Qun Liu, and Shouxun Lin.
\newblock A topic similarity model for hierarchical phrase-based translation.
\newblock In \emph{Proceedings of the 50th Annual Meeting of the Association
  for Computational Linguistics: Long Papers-Volume 1}, pp.\  750--758.
  Association for Computational Linguistics, 2012.

\bibitem[Yadan et~al.(2013)Yadan, Adams, Taigman, and Ranzato]{yadan2013multi}
Omry Yadan, Keith Adams, Yaniv Taigman, and Marc'Aurelio Ranzato.
\newblock Multi-gpu training of convnets.
\newblock \emph{arXiv preprint arXiv:1312.5853}, 2013.

\bibitem[Yu et~al.(2001)Yu, Lu, Zhu, Duan, Kang, Sun, Wang, Zhao, and
  Zhan]{yu2001processing}
Shiwen Yu, Jianming Lu, Xuefeng Zhu, Huiming Duan, Shiyong Kang, Honglin Sun,
  Hui Wang, Qiang Zhao, and Weidong Zhan.
\newblock Processing norms of modern chinese corpus.
\newblock Technical report, Technical report, 2001.

\bibitem[Zhang et~al.(2016{\natexlab{a}})Zhang, Xiong, Su, Duan, and
  Zhang]{zhang2016variational}
Biao Zhang, Deyi Xiong, Jinsong Su, Hong Duan, and Min Zhang.
\newblock Variational neural machine translation.
\newblock \emph{arXiv preprint arXiv:1605.07869}, 2016{\natexlab{a}}.

\bibitem[Zhang et~al.(2016{\natexlab{b}})Zhang, Li, Way, and
  Liu]{zhang2016topic}
Jian Zhang, Liangyou Li, Andy Way, and Qun Liu.
\newblock Topic-informed neural machine translation.
\newblock In \emph{Proceedings of COLING 2016, the 26th International
  Conference on Computational Linguistics: Technical Papers}, pp.\  1807--1817,
  2016{\natexlab{b}}.

\bibitem[Zhao \& Xing(2006)Zhao and Xing]{zhao2006bitam}
Bing Zhao and Eric~P Xing.
\newblock Bitam: Bilingual topic admixture models for word alignment.
\newblock In \emph{Proceedings of the COLING/ACL on Main conference poster
  sessions}, pp.\  969--976. Association for Computational Linguistics, 2006.

\bibitem[Zhao \& Xing(2008)Zhao and Xing]{zhao2008hm}
Bing Zhao and Eric~P Xing.
\newblock Hm-bitam: Bilingual topic exploration, word alignment, and
  translation.
\newblock In \emph{Advances in Neural Information Processing Systems}, pp.\
  1689--1696, 2008.

\bibitem[Zoph et~al.(2016)Zoph, Yuret, May, and Knight]{zoph2016transfer}
Barret Zoph, Deniz Yuret, Jonathan May, and Kevin Knight.
\newblock Transfer learning for low-resource neural machine translation.
\newblock \emph{arXiv preprint arXiv:1604.02201}, 2016.

\end{thebibliography}
\bibliographystyle{natbib}

\end{document}